\documentclass[10pt,twocolumn,letterpaper]{article}

\usepackage[pagenumbers]{cvpr} 
\usepackage[accsupp]{axessibility} 

\usepackage{booktabs}

\usepackage[most]{tcolorbox}
\usepackage{listings} 
\usepackage{xcolor}   
\definecolor{myblue}{RGB}{54,97,136} 

\definecolor{cvprblue}{rgb}{0.21,0.49,0.74}
\usepackage[pagebackref,breaklinks,colorlinks,allcolors=cvprblue]{hyperref}

\def\paperID{*****} 
\def\confName{CVPR}
\def\confYear{2026}

\title{MarkushGrapher-2: End-to-end Multimodal Recognition \\ of Chemical Structures}

\author{
  Tim Strohmeyer \textsuperscript{1, 2} \quad Lucas Morin\textsuperscript{1, 2} \quad Gerhard Ingmar Meijer\textsuperscript{1} \quad Valéry Weber\textsuperscript{1} \quad Ahmed Nassar\textsuperscript{1} \\ \quad Peter Staar\textsuperscript{1}\\
  \textsuperscript{1}IBM Research \quad \textsuperscript{2}ETH Zurich \ \\
  {\tt\small \{tis, lum, inm, vwe, ahn, taa\}@zurich.ibm.com} 
}

\begin{document}
\maketitle 

\begin{abstract}

Automatically extracting chemical structures from documents is essential for the large-scale analysis of the literature in chemistry. Automatic pipelines have been developed to recognize molecules represented either in figures or in text independently. However, methods for recognizing chemical structures from multimodal descriptions (Markush structures) lag behind in precision and cannot be used for automatic large-scale processing.
In this work, we present MarkushGrapher-2, an end-to-end approach for the multimodal recognition of chemical structures in documents. First, our method employs a dedicated OCR model to extract text from chemical images. Second, the text, image, and layout information are jointly encoded through a Vision-Text-Layout encoder and an Optical Chemical Structure Recognition vision encoder. Finally, the resulting encodings are effectively fused through a two-stage training strategy and used to auto-regressively generate a representation of the Markush structure.
To address the lack of training data, we introduce an automatic pipeline for constructing a large-scale dataset of real-world Markush structures. In addition, we present IP5-M, a large manually-annotated benchmark of real-world Markush structures, designed to advance research on this challenging task. Extensive experiments show that our approach substantially outperforms state-of-the-art models in multimodal Markush structure recognition, while maintaining strong performance in molecule structure recognition. Code, models, and datasets are released publicly.\footnote{\url{https://github.com/DS4SD/MarkushGrapher}}

\end{abstract}

\section{Introduction}

\begin{figure}[h!]
    \centering
    \includegraphics[width=0.67\textwidth,
                     trim=0mm 252mm 105mm 68mm, 
                     clip]{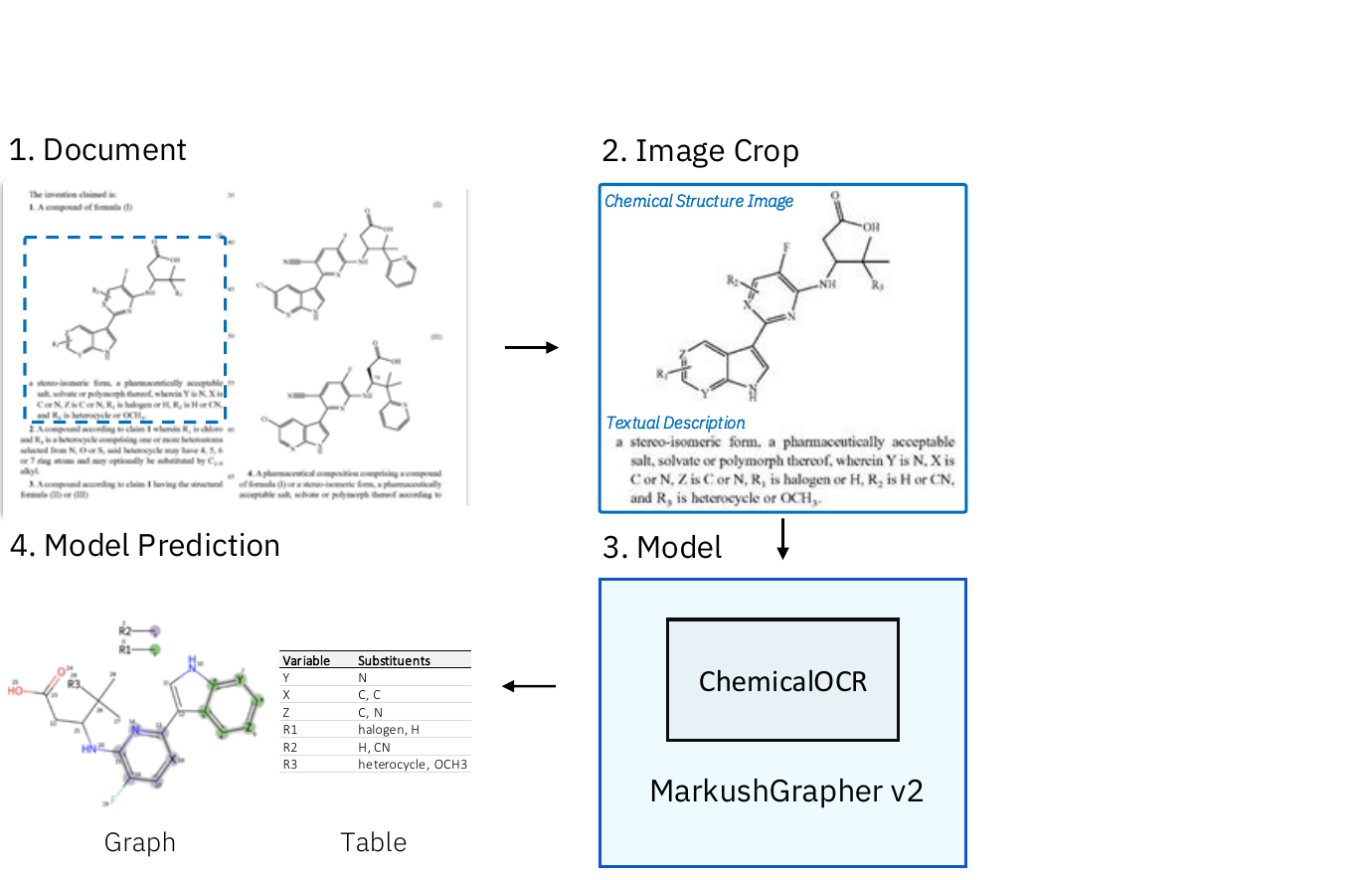}
    \caption{\textbf{Model Use Case}: MarkushGrapher-2 parses Markush backbones and variable regions from document image crops via joint multimodal encoding of vision, text, and layout.}
    \label{fig:Overview}
\end{figure}

Extracting chemical structures from documents is essential for unifying knowledge in chemistry. It enables search engines to retrieve information using molecular queries and allows the creation of training datasets for machine learning models. Converting unstructured documents into machine-readable formats can ultimately accelerate discovery across the life and materials sciences \cite{Pyzer-Knapp2025}.
Several automated pipelines have been developed to create molecular databases from unstructured documents \cite{Morin2024, Rajan2023, D5DD00313J}. These systems extract molecules either from their textual definitions, using chemical named entity recognition \cite{zhai-etal-2019-improving,doi:10.1021/acs.jcim.6b00207,Korvigo2018}, or from their visual representations, using chemical structure image segmentation \cite{Rajan2021,Zhou2023,Tang2024} and recognition \cite{Morin_2023_ICCV,Qian2023,Rajan2023}.
More recently, progress in multimodal document understanding has enabled extracting more complex chemical representations that combine visual and textual information, known as Markush structures \cite{11094170}. 
These widely used representations provide compact descriptions of families of related molecules. A Markush structure includes both an image and a text component: the image defines the Markush backbone, containing atoms, bonds, and variable regions, while the text specifies the molecular substituents that can replace those variable regions. The variable regions may include variable groups (also named residual R-groups), frequency variation indicators, and positional variation indicators \cite{doi:10.1021/ci00002a009}.
Markush structures play a central role in patent analysis, supporting prior-art searches, freedom-to-operate evaluations, and landscape analysis \cite{SIMMONS2003195}. 
Despite their importance in chemical research, their coverage in databases remains limited. Currently, Markush structures are only indexed in the proprietary manually-created databases MARPAT \cite{Ebe1991} and DWPIM \cite{doi:10.1021/ci00066a008}.

Multi-modal Markush Structure Recognition (MMSR) poses key challenges that limit the performance of their automatic recognition.
First, images of Markush structure backbones follow a wide range of conventions and drawing standards. In patent documents, for example, the visual style can vary substantially across patent offices and publication years.
Second, the textual definitions of Markush structures lack standardization and often contain condition-based or recursive descriptions.
Third, there is a lack of real-world training datasets with comprehensive annotations of Markush structure visual and textual definitions.

In this work, we introduce MarkushGrapher-2, a model for end-to-end multimodal recognition of Markush structures, illustrated in \autoref{fig:Overview}. MarkushGrapher-2 extends the MarkushGrapher \cite{11094170} framework into a universal approach capable of recognizing both molecular images and multimodal Markush structures.
The model follows an encoder–decoder architecture that takes as input an image of a molecule or Markush structure and outputs a textual sequence representing its structure. This output is divided into two components: (1) a graph of the Markush backbone, and (2) a table of possible substituents that replace the variable groups in the backbone. The input image is jointly encoded using two encoders, a Vision-Text-Layout (VTL) encoder and a vision encoder, pretrained for the task of Optical Chemical Structure Recognition (OCSR). These encodings are projected, concatenated and fed to a text decoder to autoregressively generate a sequential Markush representation.
Compared to its predecessor, MarkushGrapher-2 introduces several major improvements. First, it integrates a dedicated OCR module, enabling end-to-end processing. Here, \emph{end-to-end} refers to the model's ability to directly process a raw image at inference time, without requiring pre-annotated OCR outputs. Second, it adopts a new two-phase training strategy designed to improve the fusion of encoders. Third, the model is trained using a new training data generation pipeline, allowing it to recognize both molecule images and multi-modal Markush structures.
To further support research in this area, we introduce IP5-M, a benchmark dataset of manually annotated Markush structures extracted from patent documents of the IP5 patent offices, United States Patent and Trademark Office (USPTO),  Japan Patent Office (JPO), Korean Intellectual Property Office (KIPO), China National Intellectual Property Administration (CNIPA), and  European Patent Office (EPO). 
Comprehensive experiments across multiple benchmarks for MMSR demonstrate that MarkushGrapher-2 consistently outperforms both general-purpose and chemistry-specific document understanding models, while maintaining competitive performance on OCSR benchmarks. 
In summary:
\begin{itemize}
\item We develop MarkushGrapher-2, a universal model for recognizing both molecular images and multi-modal Markush structures.
\item We introduce a dedicated ChemicalOCR module for end-to-end processing and improved abbreviation recognition.
\item We design a two-phase training strategy to improve MarkushGrapher-2's encoder fusion.
\item We create a data generation pipeline for real-world Markush backbone training samples from MOL files and accompanying images provided by the USPTO.
\item We release IP5-M, a manually annotated benchmark of real-world Markush structures from IP5 patent offices.
\end{itemize}

\section{Related Work}

Most existing approaches to Markush structure recognition focus exclusively on the image component, neglecting supporting information through textual descriptions. Some methods, originally designed for Optical Chemical Structure Recognition (OCSR) methods, can identify a limited subset of Markush structure backbones, i.e., chemical structures containing variable groups \cite{Rajan2023, chen2024molnextrgeneralizeddeeplearning, Qian2023}. More recently, MolParser extended this capability to capture positional and frequency variation indicators \cite{Fang_2025_ICCV}. However, its handling of frequency variation is restricted to single-atom cases only. Overall, these approaches remain limited in both  scope and accuracy.
Multimodal pipelines for chemical document understanding have shown progress \cite{doi:10.1021/ci800449t,doi:10.1021/acs.jcim.4c00572}. General-purpose vision–language models have also begun to incorporate molecule-image recognition as one of their tasks, such as GOT-OCR 2.0 \cite{wei2024generalocrtheoryocr20}, Qwen2.5-VL \cite{qwen2_5vl_tech_report}, PaliGemma 2 \cite{steiner2024paligemma2familyversatile}, and DeepSeek-OCR \cite{wei2025deepseekocrcontextsopticalcompression}. Their prediction is however limited. Domain-specific VLMs like Uni-SMART \cite{cai2024unismartuniversalsciencemultimodal} and PatentFinder \cite{shi2025intelligentautomatedmolecularpatent} can also determine whether a query molecule is covered by a Markush definition in a document, which implicitly requires multimodal Markush recognition. However, these models are not practical for searching large corpuses, as each query would require running the model across all candidate documents.
MarkushGrapher \cite{11094170} is the only vision-language model capable of jointly interpreting textual and visual components of Markush definitions. Its initial version requires access to all OCR cells in the input image, which is a limitation for an integration in an end-to-end processing pipeline; and its visual recognition accuracy has room for improvement.

\section{MarkushGrapher-2}

\begin{figure*}
    \centering
    \resizebox{\textwidth}{!}{%
    \includegraphics[width=0.8\textwidth,
                     trim=7mm 75mm 15mm 50mm, 
                     clip]{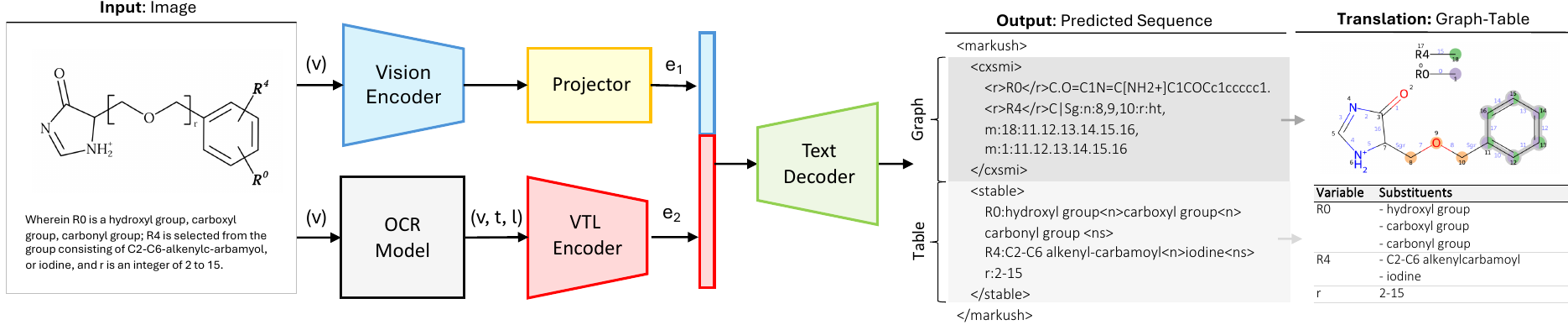}
    }
    \caption{\textbf{Model Architecture}: MarkushGraher-2 employs two complementary encoding pipelines. In the first pipeline, the input image is processed by a vision encoder (blue) followed by an MLP projector (yellow). In the second pipeline, the image is passed through an OCR model to extract textual content and bounding boxes, which is then fed into a Vision–Text–Layout (VTL) encoder together with the original image. The output of the MLP projector (e1) is concatenated with the resulting VTL embedding (e2). The combined representation is passed to a text decoder to generate a sequential description of the Markush structure and its substituents in tabular form.}
    \label{fig:model_architecture}
\end{figure*}

MarkushGrapher-2 is a transformer-based model that jointly encodes vision, text, and layout modalities for multimodal chemical structure recognition. The model integrates two complementary encoders: a Vision Encoder, pretrained for standard Optical Chemical Structure Recognition (OCSR), and a Vision–Text–Layout (VTL) Encoder, trained specifically for Markush features extraction. The latter leverages textual and positional information present in chemical documents—such as variable groups, positional and frequency variation indicators—to produce a unified multimodal representation.

\subsection{Architecture}

Figure \ref{fig:model_architecture} illustrates the model architecture, which consists of two complementary encoding pipelines. 
In the first pipeline, the input image is processed by a vision encoder (taken from MolScribe \cite{Qian2023}) pretrained for OCSR, adopting a Swin-B ViT backbone to extract visual features representing molecular structures.
In the second pipeline, the same image is passed through an OCR module that detects and recognizes textual elements within the image. These elements include atom labels, abbreviations, and descriptive text (e.g., variable group definitions) typically located near or below the structure. The extracted text and bounding boxes are combined with the image patches and fed into a VTL encoder based on a T5-base backbone. Following the UDOP fusion paradigm \cite{tang2023unifyingvisiontextlayout}, visual and textual tokens that spatially coincide—according to their bounding boxes—are aligned and fused to form a combined multimodal representation.
The output of the VTL encoder is then concatenated with a projected embedding from the pretrained Vision Encoder. This joint representation is passed to a text decoder, which autoregressively generates a structured sequence describing the chemical backbone and its associated Markush features. The generated Markush features include variable groups, frequency variation indicators, positional variation indicators, and a table containing the variable group substituents mentioned in the textual description.

\subsection{OCR Model}

The Optical Character Recognition (OCR) module is a critical component of the model architecture, providing the textual and layout modalities necessary for interpreting complex Markush features. The OCR module extracts character-level text and bounding boxes from the input image. Subsequently, the extracted text, bounding boxes and vision patches are passed to the model’s VLT encoder for multimodal encoding.

Due to limited performance of existing OCR models on chemical images, we introduce ChemicalOCR, a compact vision–language model (VLM) fine-tuned for OCR in chemical images. The ChemicalOCR architecture is based on Smoldocling~\cite{nassar2025smoldoclingultracompactvisionlanguagemodel}, a lightweight 256M-parameter model originally developed for end-to-end document conversion.

\subsection{Two-stage Training Strategy}

\begin{figure}
    \centering
    \includegraphics[width=0.67\textwidth,
                     trim=47mm 50mm 50mm 7mm, 
                     clip]{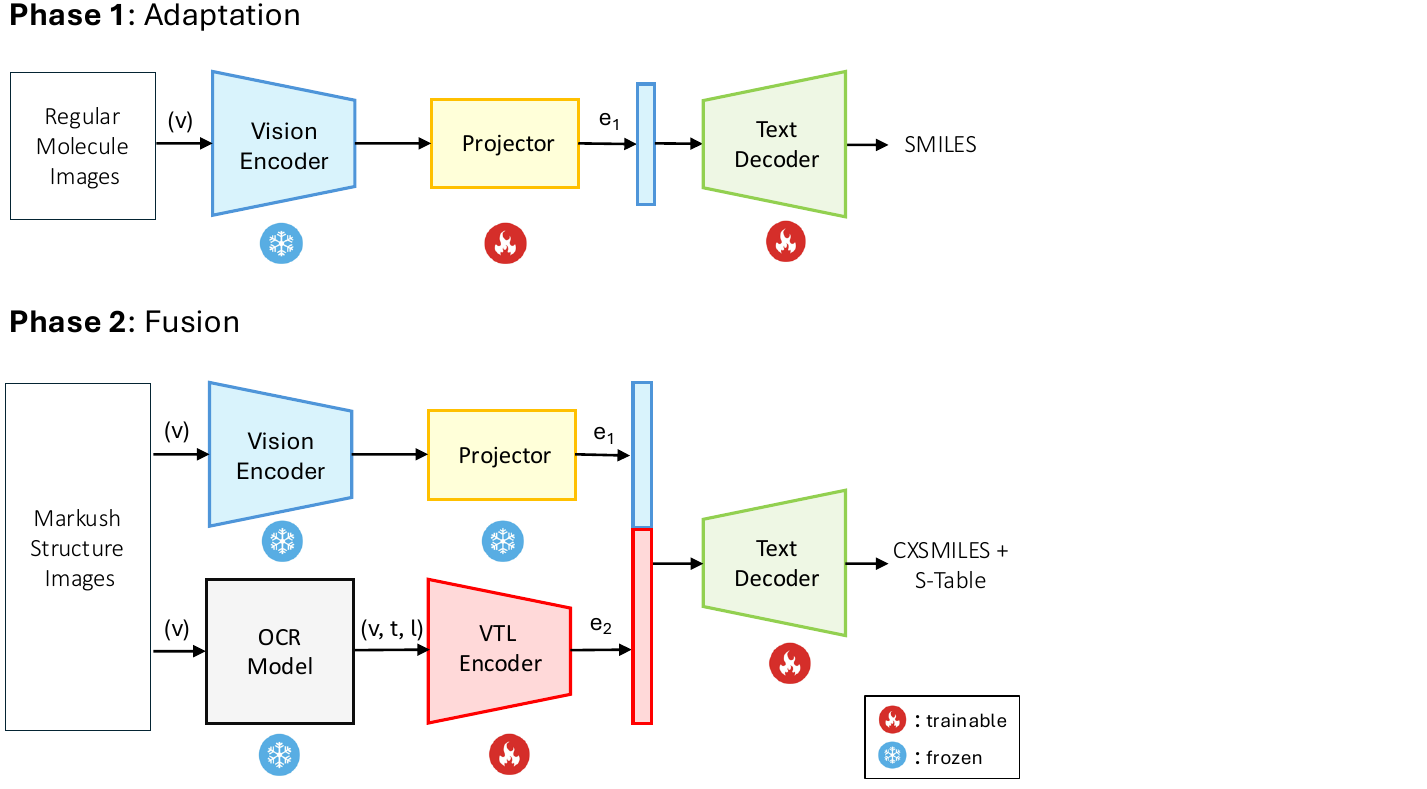}
    \caption{\textbf{Two-Phase Training}: In Phase 1 (Adaptation), the OCSR encoder is frozen while the MLP projector and text decoder are trained for SMILES prediction to align with pretrained OCSR features. In Phase 2 (Fusion), the adapted modules are initialized, the VTL encoder is introduced, and the full model is trained end-to-end for CXSMILES prediction.}
    \label{fig:two_phase_training}
\end{figure}

To train MarkushGrapher-2, we adopt a two-stage training strategy designed to fully leverage the pretrained features of the vision encoder for OCSR, while effectively fusing them with the multimodal representations learned by the VTL encoder for MMSR. Figure~\ref{fig:two_phase_training} shows a depiction of the two training phases.

During Phase 1 (Adaptation), the Vision Encoder, Projector, and Text Decoder are trained in isolation for the task of standard SMILES (Simplified Molecular Input Line Entry System) prediction. Keeping the Vision Encoder weights frozen, this allows the decoder and projector to adapt to the pretrained OCSR feature space without altering the original visual representations.
During Phase 2 (Fusion), the OCR model and VTL encoder are introduced to the model architecture. The Vision Encoder, Projector, and OCR model are frozen and the VTL encoder and Text decoder are trained end-to-end for the task of CXSMILES (Chemaxon Extended SMILES) and Substituent Table prediction. 

This two-phase training strategy allows the model to effectively leverage the pretrained Vision Encoder and build upon the OCSR features to perform the more challenging task of Markush structure recognition. By freezing the Vision Encoder and Projector during Phase 2 we preserve the original OCSR feature space and ensure that the VTL encoder focuses on learning the missing features required for Markush structure recognition (i.e., CXSMILES and substitutent table  prediction).

\section{Datasets}

MarkushGrapher-2 is trained on diverse sets of synthetic data and real-world data. The data is sourced and converted from different public document collections and databases: IP5 patent offices, PubChem, and datasets published by MolScribe \cite{Qian2023} and MolParser \cite{fang2025molparserendtoendvisualrecognition}.

\subsection{The USPTO-MOL-M Dataset}

For the training of MarkushGrapher-2, we introduce a new large-scale training dataset, consisting of real-world image-(CX)SMILES pairs extracted automatically from MOL files that are provided by the USPTO along with the patent document. The USPTO MOL files (V2000 format) contain information such as atom symbols, atomic coordinates, bonds and bond types, aliases (abbreviations and R-groups), superatoms (abbreviations), and frequency variation indicators. These MOL files do not include all the visual details that are shown in their corresponding image, for example, positional variation indicators are not directly included. We developed code to clean the MOL files and convert them to CXSMILES format. In this process, positional variation indicators were reconstructed based on atomic coordinates and bonding patterns, frequency variation indicators were extracted from the structural repeating unit sections, and R-groups were derived from the alias and superatom sections. Unnecessary elements—such as text, labels, or indices present in the MOL files—were removed to the greatest extent possible. The dataset built in this study comprises USPTO MOL files from the years 2010 to 2025.

\subsection{Training Datasets}
\label{training_dataset}

The datasets used for training MarkushGrapher-2 can be categorized into three categories: (1) OCR data, consisting of image-OCR pairs for optical character recognition on chemical structure images, (2) OCSR data, consisting of image–SMILES pairs for standard molecular structure recognition, and (3) MMSR data, comprising image–CXSMILES pairs (and substituent tables) for recognizing Markush structures.

ChemicalOCR module is pretrained on a set of 235k synthetic chemical structures containing automatic OCR annotations. To generate the training images, SMILES are randomly sampled from PubChem \cite{Kim2025-tc} and augmented into chemically valid CXSMILES representations, introducing variable groups, frequency and positional variation indicators. These CXSMILES are then rendered into images containing corresponding molecular drawings, textual annotations, and character-level bounding boxes \cite{11094170}. To substantially improve model performance on real-world chemical images, the model is further finetuned using a set of 7k manually annotated OCR samples with chemical structures cropped from IP5 patent documents.

During Phase 1 (Adaptation), the model is trained exclusively on OCSR data. Specifically, we use 243k real-world image–SMILES pairs sourced from the public MolScribe dataset \cite{Qian2023}, derived from USPTO documents.

In the subsequent Phase 2 (Fusion), we train the model on a combination of MMSR datasets. This includes same 235k synthetically generated image–CXSMILES pairs with corresponding substituent tables that were used for training the ChemicalOCR module. Additionally, the MMSR training corpus includes 91k real-world Markush samples from the MolParser dataset \cite{fang2025molparserendtoendvisualrecognition}, that are converted into an optimized CXSMILES prediction format, and 54k real-world Markush samples from the USPTO-MOL-M dataset described above.

\section{Experiments}


\subsection{Implementation Details}

The Vision-Text-Layout encoder and text decoder are based on a T5 encoder–decoder architecture \cite{tang2023unifyingvisiontextlayout}. The OCSR encoder is the vision encoder taken from MolScribe, which is based on a SWIN transformer architecture \cite{Qian2023}. This encoder remains frozen during training. In total, the model comprises 831M parameters, of which 744M are trainable. Training is performed in two phases. During Phase 1 (Adaptation), the Projector and Text Decoder are trained on 243K real samples for 3 epochs using an NVIDIA A100 GPU. Here, we use the Adam optimizer with a learning rate of 5e-4, 1000 warm-up steps, a batch size of 10, and a weight decay of 1e-3. For Phase 2 (Fusion), the pretrained Vision encoder and Projector are frozen, and the VTL encoder and text decoder are further trained using a mix of 235k synthetic and 145k real-world Markush structure samples. Here we use a batch size of 8 and train for 2 epochs.

\subsection{Evaluation Datasets and Metrics}

We evaluate MarkushGrapher-2 on two tasks: (1) image recognition and (2) substituent table recognition.

\textbf{Markush Benchmarks:} To evaluate model performance for Markush structure recognition under diverse real-world conditions, we employ several benchmark datasets of Markush structures. M2S contains 103 real-world Markush structure images with textual descriptions, manually annotated and substituent tables \cite{11094170}. USPTO-M contains 74 real-world Markush structure images, manually annotated \cite{11094170}. WildMol-M, introduced by MolParser \cite{fang2025molparserendtoendvisualrecognition}, contains 10k real-world, semi-manually annotated Markush structure images. Finally, we introduce a new benchmark, IP5-M, consisting of 1,000 manually annotated Markush structures from patent documents of the IP5 offices published between 1980 and 2025.

\textbf{OCSR Benchmarks:}
To evaluate model performance on standard SMILES prediction, we use four common OCSR benchmarks, USPTO \cite{Filippov2009-ag}, JPO \cite{Fujiyoshi2011}, UOB \cite{Sadawi2012}, and WildMol \cite{fang2025molparserendtoendvisualrecognition}.

\textbf{OCR Benchmarks:} The above M2S, USPTO-M, and IP5-M benchmarks are also used to evaluate the ChemicalOCR performance on OCR predictions.

\textbf{CXSMILES Accuracy (A):} Image recognition performance is measured using the CXSMILES accuracy (A), while substituent table recognition is evaluated using prediction accuracy and F1 score. Markush accuracy measures the overall accuracy of correct Table and CXSMILES predictions. Stereochemistry is ignored during evaluation to ensure consistency across datasets. In more detail, the CXSMILES accuracy (A) metric quantifies the percentage of perfectly recognized molecular representations. A prediction is considered correct if two conditions are satisfied: (1) the predicted SMILES (without Markush features) corresponds to the ground truth according to \texttt{InChIKey} equivalence, and (2) the Markush features, i.e., variable groups (R-groups), positional and frequency variation indicators are correctly represented.

\subsection{Evaluation and State-of-the-art Comparison}

\begin{table*}
\centering
\resizebox{\textwidth}{!}{%
\begin{tabular}{lcccccccccccccccc}
\toprule
\textbf{Models} & \multicolumn{4}{c}{\textbf{M2S (103)}} & \multicolumn{4}{c}{\textbf{USPTO-M (74)}}  & \multicolumn{4}{c}{\textbf{IP5-M (1000)}}\\
\cmidrule(lr){2-5} \cmidrule(lr){6-9} \cmidrule(lr){10-13} 
 & P & R & F1 & A@IoU\_0.5 & P & R & F1 & A@IoU\_0.5 & P & R & F1 & A@IoU\_0.5 \\
\midrule
PaddleOCR v5    & 8.9 & 6.8 & 7.7 & 0.0 & 2.3 & 1.1 & 1.2 & 0.0 & 2.2 & 1.7  & 1.9 & 0.6 \\
EasyOCR        & 9.8 & 10.7 & 10.2 & 0.0 & 24.8 & 14.2 & 18.0 & 0.0 & 23.5 & 15.2 & 18.4 & 2.7 \\
\textbf{ChemicalOCR (Ours)}     & \textbf{86.9} & \textbf{87.4} & \textbf{87.2} & \textbf{32.0} & \textbf{93.5} & \textbf{92.6} & \textbf{93.0} & \textbf{63.5} & \textbf{85.6} & \textbf{87.4} & \textbf{86.5} & \textbf{69.5} \\
\bottomrule
\end{tabular}
}
\vspace{1em} 
\caption{\textbf{OCR on Chemical Images}: Comparison of our ChemicalOCR model with existing OCR models. The evaluation is conducted on real-world benchmarks (M2S, USPTO-M, and IP5-M). Precision $P$, Recall $R$, and F1 are measured at individual bounding-box level. Accuracy $A$ is measured at the image level; an image is considered correct if all OCR cells have an  IoU $>0.5$ and their recognized characters match the ground truth.
}\label{ocr_scores}
\end{table*}

\begin{figure*}
    \centering
    \includegraphics[width=0.81\textwidth,
                     trim=2mm 273mm 0mm 0mm, 
                     clip]{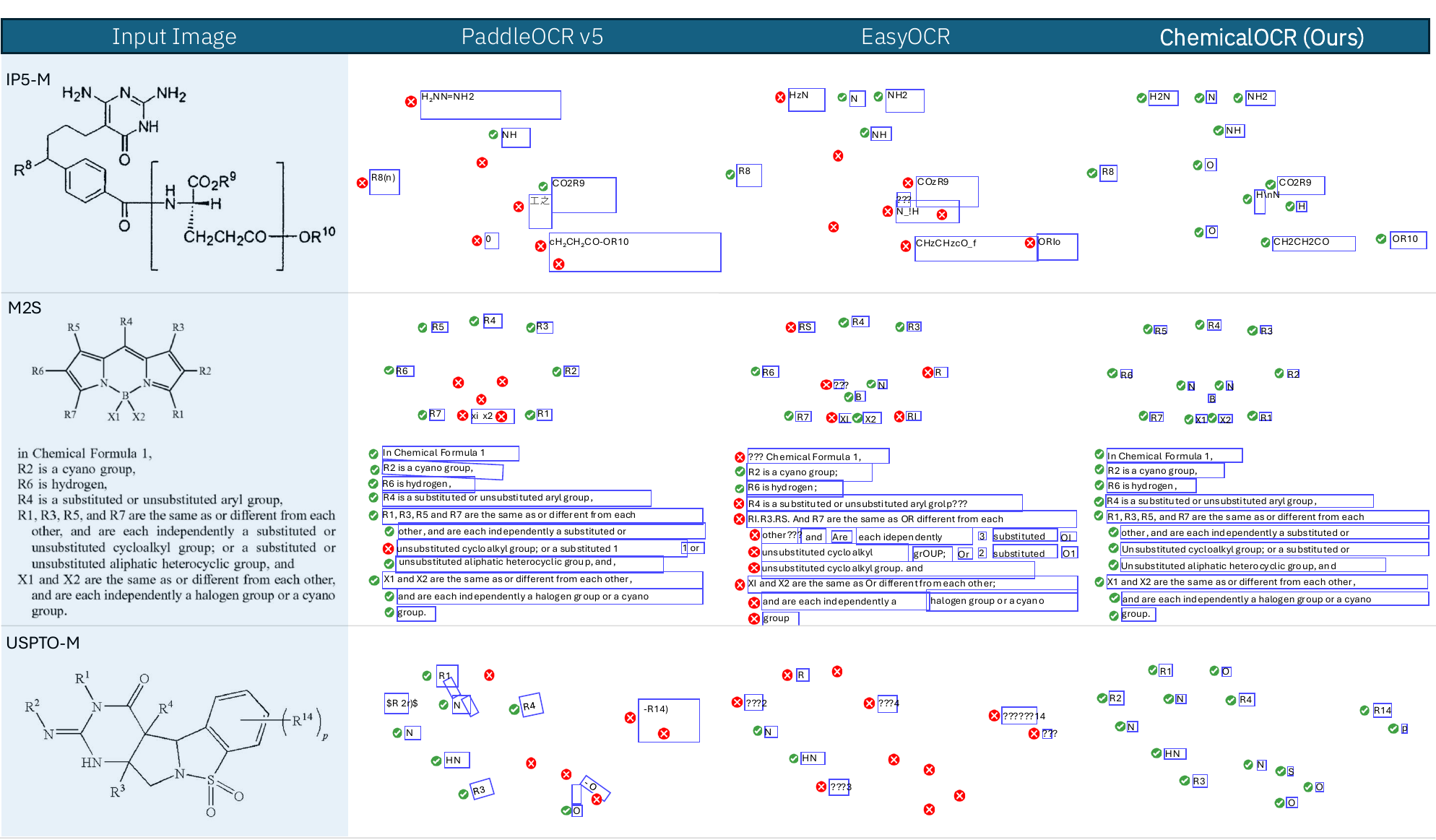}
    \caption{\textbf{OCR - Qualitative Comparison}: Comparison of OCR predictions by three models PaddleOCR v5, EasyOCR, and ChemicalOCR (Ours) for an exemplary chemical structure from the benchmarks M2S, USPTO-M, and IP5-M. Red labels indicate incorrect OCR, green labels indicate correct OCR and blue indicates predicted bounding boxes.}
    \label{fig:ocr_predictions}
\end{figure*}

\begin{table*}
\centering
\resizebox{\textwidth}{!}{%
\begin{tabular}{lccccccccccc}
\toprule
\textbf{Methods} &
\multicolumn{4}{c}{\textbf{M2S (103)}} &
\textbf{USPTO-M (74)} &
\textbf{WildMol-M (10000)} &
\textbf{IP5-M (1000)}\\
 \cmidrule(lr){2-5}  
 & \textbf{CXSMILES} & \multicolumn{2}{c}{\textbf{Table}} & \textbf{Markush} 
 & \textbf{CXSMILES} & \textbf{CXSMILES} & \textbf{CXSMILES} \\
 & A & A & F1 & A & A & A & A \\
\midrule
\textit{Image only} \\
MolParser-Base     & 39 & -- & -- & -- & 30 & 38.1\textsuperscript{$\ddagger$} & 47.7\\
MolScribe          & 21\textsuperscript{$\dagger$} & -- & -- & -- & 7\textsuperscript{$\dagger$} & 28.1 & 22.3 \\
\midrule
\textit{Multimodal} \\
GPT-5              &  3  & 8  & 24  & 0  &   &   \\
DeepSeek-OCR       & 0  & --  & --  & --  & 0  & 1.9 & 0.0\\
MarkushGrapher-1  & 38\textsuperscript{$\dagger$} & \textbf{29}\textsuperscript{$\dagger$} & \textbf{65}\textsuperscript{$\dagger$} & 10\textsuperscript{$\dagger$} & 32\textsuperscript{$\dagger$} & -- & --\\
\textbf{MarkushGrapher-2 (Ours)}  & \textbf{56} & 22 & \textbf{65} & \textbf{13} & \textbf{55} & \textbf{48.0} & \textbf{53.7}\\
\bottomrule
\end{tabular}
}
\vspace{0.5em}
\caption{\textbf{Markush Structure Recognition:} Comparison of our MarkushGrapher-2 model with existing models on CXSMILES and Substituent Table prediction. Models are evaluated on real-world benchmarks (M2S, UPSTO-M, WildMol-M, IP5-M). Accuracy $A$ measures the percentage of correctly predicted samples, while F1 quantifies the similarity between predictions and ground truth, ranging from 0 (least similar) to 100 (most similar). 
\textsuperscript{$\dagger$} scores taken from  \cite{11094170}, \textsuperscript{$\ddagger$} scores taken from \cite{fang2025molparserendtoendvisualrecognition}.}
\label{tab:mmsr_results}
\end{table*}

In this section we provide an evaluation of the ChemicalOCR module and the overall MarkushGrapher-2 performance versus state-of-the-art methods. 

\textbf{Optical Character Recognition:} Table \ref{ocr_scores} shows a quantitative comparison of our ChemicalOCR model with PaddleOCR v5 \cite{cui2025paddleocr30technicalreport} and EasyOCR \cite{easyocr}. ChemicalOCR substantially outperforms PaddleOCR and EasyOCR. Figure \ref{fig:ocr_predictions} presents a visual comparison of OCR predictions obtained by each evaluated method for a representative Markush structure image from each benchmark. Consistent with the quantitative results in Table \ref{ocr_scores}, ChemicalOCR demonstrates superior performance in both character localization and recognition. It accurately identifies long abbreviations within chemical structures and correctly parses textual descriptions below the images (see M2S, second row). It is observed that EasyOCR struggles with longer text sequences, including abbreviations and descriptive text. PaddleOCR v5 handles longer text sequences, such as descriptive captions, reliably, but often misinterprets characters within chemical structures—frequently merging symbols into a single bounding box and confusing bonds with minus or equal signs.

\begin{table*}
\centering
\begin{tabular}{lcccc}
\toprule
\textbf{Methods}  & \textbf{WildMol (10000)} & \textbf{JPO (450)} & \textbf{UOB (5740)} & \textbf{USPTO (5719)} \\
\midrule
\textit{Image only} \\
MolParser-Base     & \textbf{76.9} & \textbf{78.9} & 91.8 & \underline{93.0}  \\
MolScribe          & 66.4 & \underline{76.2} & 87.4 & \textbf{93.1} \\
DECIMER 2.7        & 56.0 & 64.0  & 88.3  & 59.9   \\
MolGrapher       & 45.5 & 67.5 & \underline{94.9} & 91.5 \\
\midrule
\textit{Multi-modal} \\
GPT-5              &      & 19.2 &   &   \\
DeepSeek-OCR       &  25.8    &   31.6   &  78.7    &   36.9   \\
MarkushGrapher-1  & - & - & - &  -  \\
\textbf{MarkushGrapher-2 (Ours)} & \underline{68.4} & 71.0  & \textbf{96.6}   &  89.8  \\
\bottomrule
\end{tabular}%
\vspace{1em} 
\caption{\textbf{Molecular Structure Recognition}: Comparison of our MarkushGrapher-2 model with existing models on SMILES prediction. Models are evaluated on real-world benchmarks (WildMol, JPO, UOB, and USPTO). Accuracy measures the percentage of correctly predicted molecular structure. Scores for MolParser, MolScribe, DECIMER, and MolGrapher scores are taken from \cite{11094170, fang2025molparserendtoendvisualrecognition}.}
\label{tab:ocsr_results}
\end{table*}

\begin{figure*}
    \centering
    \includegraphics[width=0.71\textwidth,
                     trim=0mm 145mm 0mm 0mm, 
                     clip]{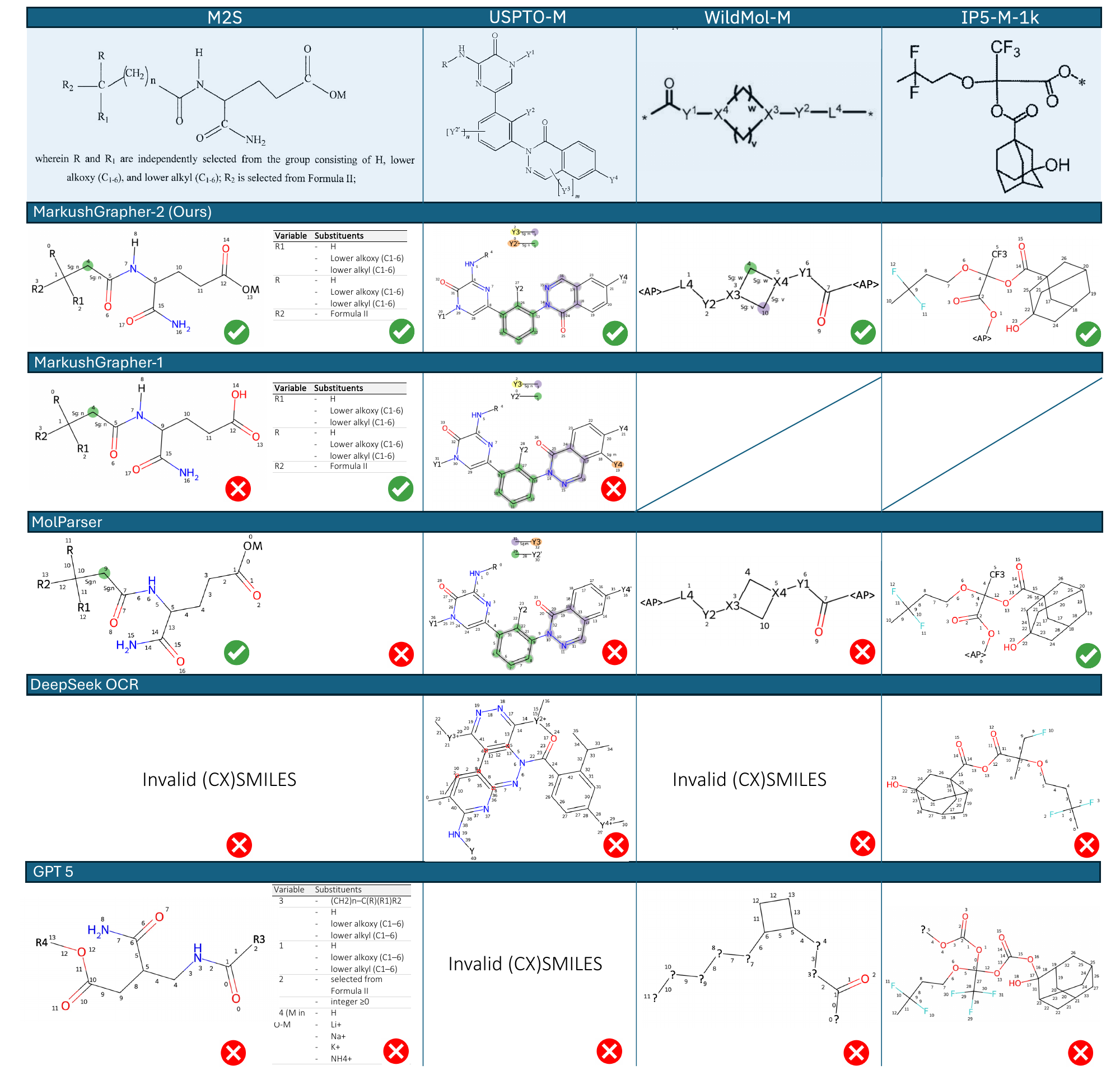}
    \caption{\textbf{Markush Structure Recognition - Qualitative Comparison}: Comparison of Markush structure predictions by five models MarkushGrapher-2 (Ours), MarkushGrapher-1, MolParser, DeepSeek OCR, and GPT-5 for an exemplary Markush structure from the benchmarks M2S, USPTO-M, WildMol-M, and IP5-M. Red labels indicate incorrect predictions,
    green labels indicate correct predictions.}
    \label{fig:cxsmiles_predictions}
\end{figure*}

\textbf{Markush Structure Recognition}: Table \ref{tab:mmsr_results} shows a quantitative comparison of our MarkushGrapher-2 model with image-only models MolParser-Base and MolScribe, and multimodel models GPT-5, DeepSeek-OCR, and MarkushGrapher-1. GPT-5 was run on the M2S benchmark, which is reasonably representative for Markush structure recognition. MarkushGrapher-2 substantially outperforms state-of-the-art models on Markush structure recognition. Figure \ref{fig:cxsmiles_predictions} presents a visual comparison of predictions of MarkushGrapher-2 and state-of-the-art models for a representative Markush structure image from each benchmarks. As illustrated, MarkushGrapher-2 accurately reconstructs both the molecular backbone from the structural image, as well as complex Markush-specific features, such as variable groups, positional and frequency variation indicators. Specifically, the model effectively captures any-locant cycle connections, repeating structural units, and end-to-molecule attach points (AP). Note: Scores for MarkushGrapher-1 cannot be provided for WildMold-M, since MarkushGrapher-1 does not constitute an OCR module.

\textbf{Molecular Structure Recognition}: Table \ref{tab:ocsr_results} shows a quantitative comparison of our MarkushGrapher-2 model with image-only models MolParser-Base and MolScribe, and multimodel models GPT-5, DeepSeek-OCR, and MarkushGrapher-1. GPT-5 was run on the JPO benchmark, which is reasonably representative for molecular structure recognition. The results show that MarkushGrapher-2 is competitive with state-of the-art models on molecular structure recognition.

\subsection{Ablation Study}

\begin{figure}
    \centering
    \includegraphics[width=0.28\textwidth, 
                     trim=0mm 260mm 160mm 0mm, 
                     clip]{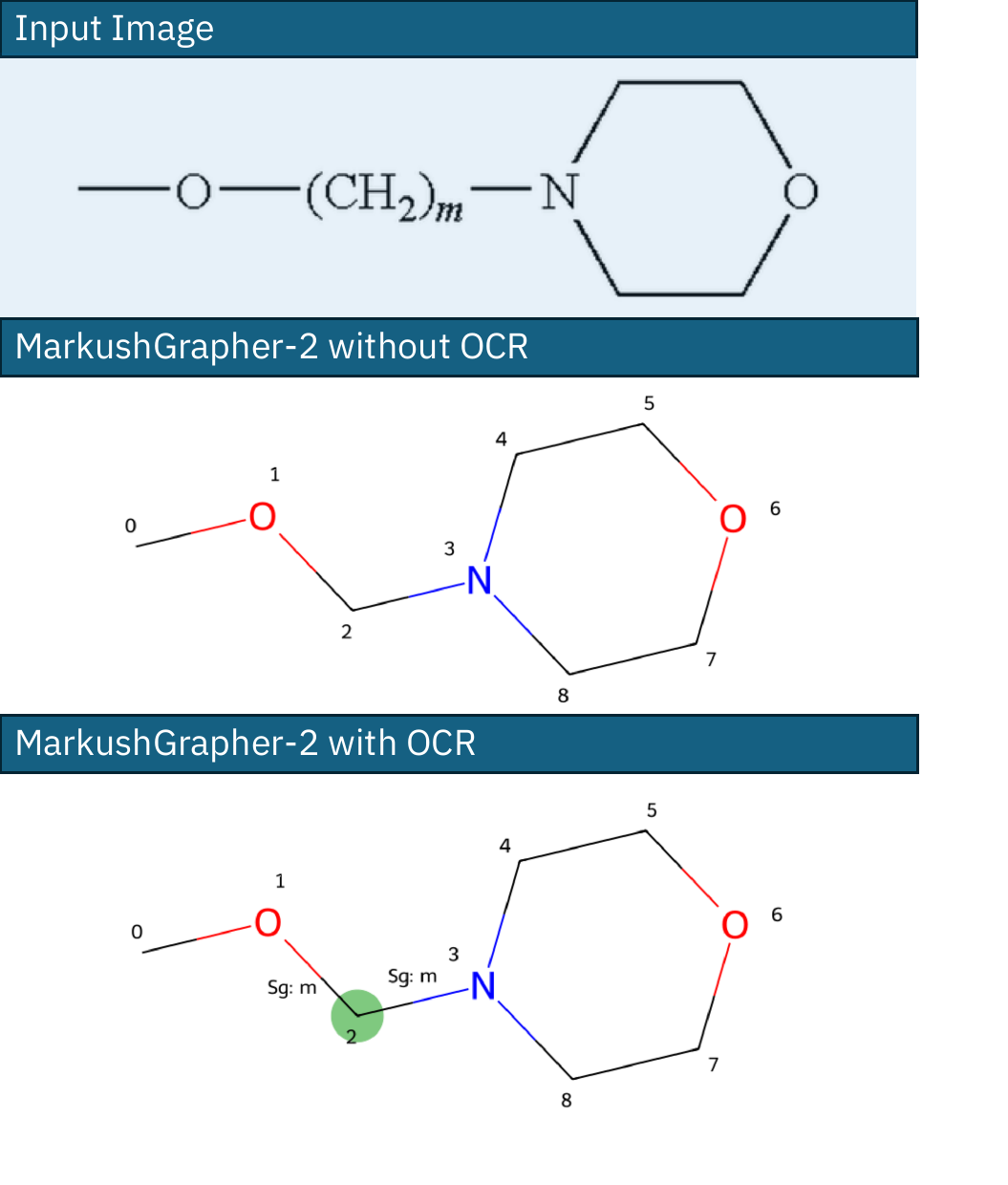}
    \caption{\textbf{Effect of OCR Predictions}: Comparison of MarkushGrapher-2 predictions with and without OCR input. The green circle indicates prediction of a frequency variation indicator (i.e., repeating Sg groups).}
    \label{fig:without_ocr_visual}
\end{figure}

\textbf{Effect of OCR module}: Table \ref{table: with_without_ocr} compares the performance of MarkushGrapher-2 with and without OCR predictions from the ChemicalOCR module as input. The scores provide insight about the importance of the text and layout modality for the overall model prediction. We observe that the OCR predictions substantially improves MarkushGrapher-2 prediction accuracy. Figure \ref{fig:without_ocr_visual} illustrates an example prediction from MarkushGrapher-2 with and without OCR input. While MarkushGrapher-2 without OCR input accurately predicts the structural backbone, it fails to capture the Markush features, i.e., the repeating groups (Sg). It may be inferred that the text extracted by the OCR—such as brackets and indices—provides important additional information that substantially improves Markush feature prediction.

\begin{table}
\centering
\resizebox{\columnwidth}{!}{%
\begin{tabular}{lcccccc}
\toprule
\textbf{Methods} & \multicolumn{2}{c}{\textbf{M2S}} & \multicolumn{2}{c}{\textbf{USPTO-M}} & \multicolumn{2}{c}{\textbf{IP5-M}} \\
\cmidrule(lr){2-3} \cmidrule(lr){4-5} \cmidrule(lr){6-7}
 & A & A$_{\text{InChIKey}}$ & A & A$_{\text{InChIKey}}$ & A & A$_{\text{InChIKey}}$ \\
\midrule
\textbf{without OCR} & 4 & 39 & 3 & 51 & 15.4 & 51.3  \\
\textbf{with OCR}    & 56 & 80 & 55 & 69 & 53.7 & 73.3 \\
\bottomrule
\end{tabular}%
}
\vspace{0.5em}
\caption{\textbf{Effect of ChemicalOCR on Overall Model Performance}: Comparison of MarkushGrapher-2 performance with and without OCR predictions as input. The table shows the CXSMILES prediction accuracy $A$ and structural backbone prediction accuracy $A_{\text{InChIKey}}$.}
\label{table: with_without_ocr}
\end{table}

\textbf{Effect of Two-Phase Training}: Table \ref{table: ablation_two_phase} shows the impact of the proposed two-phase training on overall MarkushGrapher-2 performance. We compare the two-phase setup—first adapting the Projector and VTL decoder to the pretrained OCSR features, followed by joint training of both encoders—with a single-phase training approach. In the table, the single-phase and two-phase setups are labeled with ‘Fusion only’ and ‘Adaptation and Fusion’, respectively. This experiment highlights the benefits of adapting the decoder to the pretrained OCSR features before fusing encoder outputs for Markush structure recognition. The results show that two-phase training improves the model’s ability to encode Markush features while preserving performance on standard molecular recognition.

\begin{table}
\centering
\resizebox{\columnwidth}{!}{%
\begin{tabular}{lcccccc}
\toprule
\textbf{Methods} & \multicolumn{2}{c}{\textbf{M2S}} & \multicolumn{2}{c}{\textbf{JPO}} \\
\cmidrule(lr){2-3} \cmidrule(lr){4-5}
 & A & A$_{\text{InChIKey}}$ & A & A$_{\text{InChIKey}}$ \\
\midrule
\textbf{Fusion only}               & 44 & 53 & 53.0 & 53.0 \\
\textbf{Adaptation and Fusion}    & 50 & 68 & 61.5 & 61.5 \\
\bottomrule
\end{tabular}%
}
\vspace{0.5em}
\caption{\textbf{Effect of Two-Phase Training}: Comparison of MarkushGrapher-2 performance with and without two-phase training. ‘Fusion only’ denotes the model trained in a single phase, while ‘Adaptation and Fusion’ refers to the two-phase training setup. Scores for CXSMILES prediction accuracy A and
structural backbone prediction accuracy A$_{\text{InChIKey}}$ are reported after 2 epochs for both configurations. }
\label{table: ablation_two_phase}
\end{table}

\section{Conclusion}

In this work, we present MarkushGrapher-2, a universal model for the recognition of both molecular structures and multi-modal Markush structures. Our approach introduces a dedicated ChemicalOCR module that extracts text from images, allowing the joint encoding of image, text, and layout modalities for end-to-end processing. To train MarkushGrapher-2, we employ a two-phase training strategy  designed to optimally leverage pretrained features from a vision encoder, effectively fusing them for improved Markush structure recognition. 
To address the scarcity of training data, we developed a data generation pipeline for constructing real-world Markush samples from MOL files and accompanying USPTO images. In addition, we introduce IP5-M, a manually annotated benchmark dataset of real-world Markush structures from IP5 patent documents.
Extensive experiments demonstrate that MarkushGrapher-2 substantially outperforms state-of-the-art models on Markush structure recognition, while remaining competitive on standard molecular structure recognition tasks. Our work bridges the gap between molecular and Markush recognition, offering a unified solution for large-scale automated extraction of chemical structures in documents.

\clearpage
\setcounter{page}{1}
\maketitlesupplementary

%
In this section we provide supplementary information on the work presented in this paper. This includes details on the CXSMILES schema, training and benchmark datasets, OCR training, evaluation, and multi-modal model predictions.

\section{CXSMILES - Schema}

\begin{figure}[!h]
    \centering
    \includegraphics[width=0.45\textwidth,
                     trim=0mm 330mm 168mm 10mm, 
                     clip]{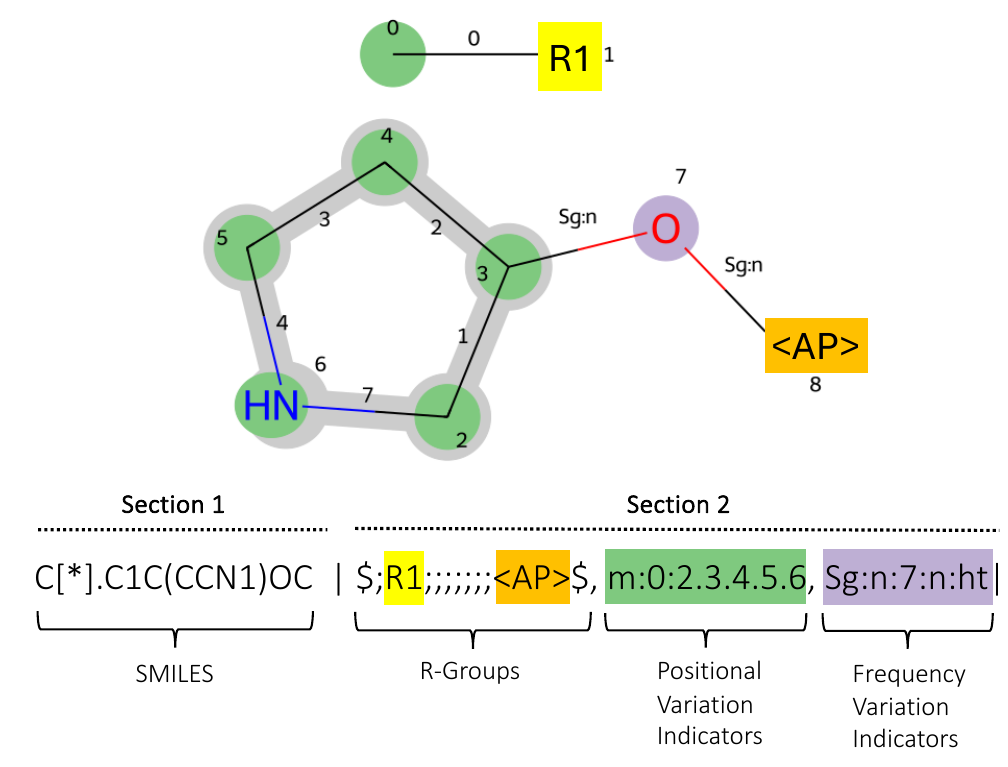}
    \caption{\textbf{CXSMILES - Schema}: A CXSMILES consists of two sections separated by a $|$ character. The first section is a SMILES that describes the molecular backbone of the Markush structure. The second section are the Markush features; variable groups (yellow) and attachment points (orange) are defined in the the R-groups section, cycle connections (green) are defined in the positional variation indicator section, and repeating structural groups (purple) are defined in the frequency variation indicator section.}
    \label{fig:cxsmiles_schema}
\end{figure}

Figure \ref{fig:cxsmiles_schema} illustrates the CXSMILES schema used to represent Markush structures in string format. The schema consists of two sections: (1) SMILES, (2) Markush features. The SMILES is a string representing the molecular structure comprising atom labels and bonds \cite{weininger_smiles_1988}. The Markush features are R-groups (variable groups and attach points), positional variation indicators, and frequency variation indicators. More details on the CXSMILES schema can be found in \cite{Chemaxon_CXSMILES}.

\section{USPTO-MOL-M Dataset}

\begin{figure}
    \centering
    \includegraphics[width=0.67\textwidth,
                     trim=6mm 50mm 134mm 2mm, 
                     clip]{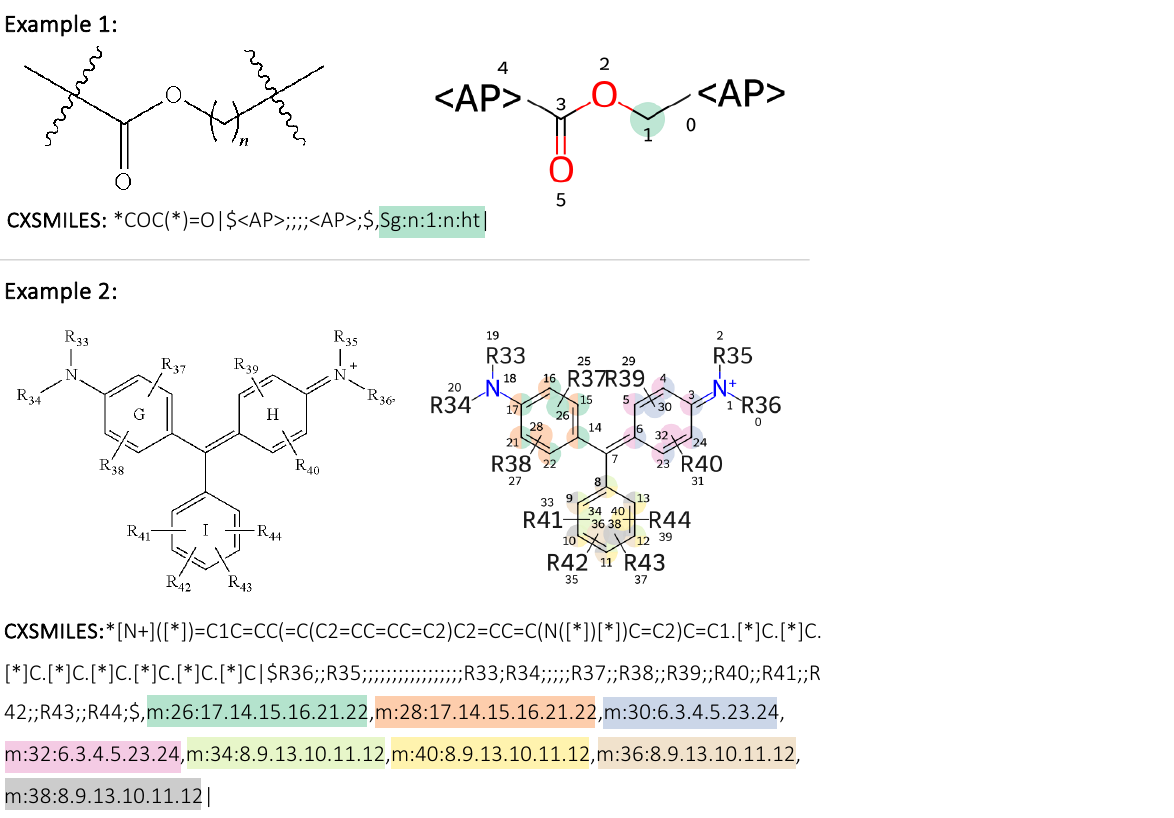}
    \caption{\textbf{USPTO-MOL-M Training dataset}: Two examples of a chemical-structure image that is provided by the USPTO (left), the CXSMILES that we obtained (bottom) by processing the corresponding MOL file that is provided by the USPTO, and a visualization of said CXSMILES (right). Example 1 shows a simple Markush structure. Typical features of Markush structures, attach point (AP) and frequency variation "n" (marked in green in CXSMILES and visualization (right)), are indicated. Example 2 shows a typical Markush structure. Typical features of Markush structures, variable groups and positional variation (color coded in CXSMILES and visualization (right)), are indicated.}
    \label{fig:uspto_example}
\end{figure}

Figure \ref{fig:uspto_example} presents two representative examples from the USPTO-MOL-M training dataset. The dataset contains real images sourced directly from USPTO patents; the corresponding CXSMILES are generated using MOL files.

\noindent Example 1 shows a simple Markush structure containing a frequency variation indicator (denoted by the brackets and index n) and an attach point. Example 2 depicts a typical Markush structure, containing multiple variable groups and positional variation indicators. Table \ref{fig:uspto_mol_m_bechmark_statistics} shows a breakdown of the Markush features present in the USPTO-MOL-M dataset.

\begin{table}
\centering
\resizebox{0.441\columnwidth}{!}{%
\begin{tabular}{lc}
\toprule
\multicolumn{2}{c}{\textbf{USPTO-MOL-M (54k)}} \\
\midrule
Variable group  & 91 \\
Attach point  & 8 \\
m-section & 42 \\
Sg-section & 55 \\
\midrule
Mean num. atoms & 20 \\
Mean num. OCR cells & 13 \\
\bottomrule
\end{tabular}
}
\caption{\textbf{USPTO-MOL-M - Statistics}: Percentage of different (Markush) features found in the training dataset.}
\label{fig:uspto_mol_m_bechmark_statistics}
\end{table}



\section{Benchmark Datasets}

\begin{figure}[h]
    \centering
    \includegraphics[width=0.555\textwidth,
                     trim=2mm 430mm 50mm 14mm, 
                     clip]{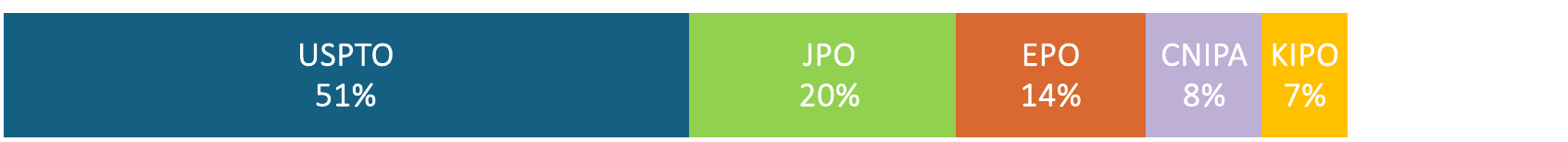}
    \caption{\textbf{IP5-M Benchmark - Composition}: Distribution of the Markush structure images extracted from patent documents of the IP5 offices.}
    \label{fig:abblation_composition}
\end{figure}

The IP5 patent offices are arguably the most relevant patent offices, handling about 90\% of the world's patent documents \cite{IP5}. Therefore, a Markush structure recognition model is only relevant if it performs well on Markush structures images in patent documents from these patent offices. The IP5-M benchmark contains 1000 such Markush structure images. Figure \ref{fig:abblation_composition} illustrates the dataset distribution. Table \ref{abblation_bechmark_statistics} summarizes key statistics for all the Markush structure recognition benchmarks used in this work.

\section{Ablation - Architectural Branches}

\autoref{tab:architectural_branches_abblation} provides an ablation of the architectural branches, comparing each encoding pipeline when trained individually with the decoder. The results show that Pipeline 1 (vision encoder + projector + decoder) achieves strong performance on molecular structure recognition (USPTO), and struggles with multimodal Markush structure recognition (M2S). In contrast, Pipeline 2 (OCR + VTL encoder + decoder) performs decently on Markush benchmarks while yielding weaker result on OCSR.
This ablation demonstrates our architecture's ability to successfully integrate the complementary strengths of both pipelines through the two-phase training. 

\begin{table}[h]
\centering
\resizebox{1.0\linewidth}{!}{
\begin{tabular}{lccc}
\hline
Methods & \multicolumn{2}{c}{M2S} & \multicolumn{1}{c}{USPTO}  \\ \cline{2-3} \cline{4-4}
& \multicolumn{1}{c}{CXSMILES} & \multicolumn{1}{c}{Table}  & \multicolumn{1}{c}{SMILES}  \\ \hline
Pipeline 1  & 8 & 0  & \underline{89.1}   \\ 
Pipeline 2 & \underline{39} & \underline{21} & 84.4 \\ 
Pipeline 1 \& 2 & \textbf{56} &  \textbf{22} &  \textbf{89.8} \\ \hline
\end{tabular}}
\caption{\textbf{Architectural Branches Ablation}: Comparison of Pipelines 1, 2, and 1 \& 2. Pipeline 1: vision encoder + projector + decoder; Pipeline 2: OCR + VTL encoder + decoder.}
\label{tab:architectural_branches_abblation}
\end{table}

\section{ChemicalOCR - Training}

ChemicalOCR is trained in two stages. In the first stage, the model is pretrained on 235k synthetically generated images containing automatic OCR annotations. In the second stage, the model is finetuned on 7k images from IP5 patent documents containing manual OCR annotations. Table \ref{ablation_ocr_training_stages} shows a comparison of ChemicalOCR's prediction accuracy after each stage. The results demonstrate that finetuning on a small set of real-world data provides a substantial improvement of OCR prediction accuracy.

\begin{table}[h]
\centering
\resizebox{\columnwidth}{!}{%
\begin{tabular}{lccc}
\toprule
\textbf{Checkpoints} & \textbf{M2S} & \textbf{USPTO-M} & \textbf{IP5-M}\\
\midrule
\textbf{ChemicalOCR - pretrained}    & 26.2 & 18.9 & 27.6 \\
\textbf{ChemicalOCR - finetuned}     & 32.0 & 63.5 & 69.5 \\
\bottomrule
\end{tabular}
}
\caption{\textbf{Comparison of ChemicalOCR training stages}: Comparison of OCR Accuracy A at IoU $>$ 0.5 of ChemicalOCR at different training stages. “Pretrained” denotes training on synthetic data only, while “finetuned” indicates additional training on 7k real samples.}
\label{ablation_ocr_training_stages}
\end{table}

\section{OCR - Benchmark Evaluations}

Table \ref{abblation_ocr_scores} shows the OCR prediction scores for different IoU thresholds. The IoU threshold determines the overlap between the predicted and ground truth bounding boxes. Possibly surprising, we observe a significant degradation in PaddleOCR’s accuracy as the IoU requirement increases. This trend aligns with the qualitative results displayed in Figure 4 of the main paper. 
We note that PaddleOCR tends to misinterpret chemical features, such as bonds, as characters. This leads to the prediction of large, incorrect bounding boxes. In comparison, ChemicalOCR and EasyOCR provide more consistent results for different IoU thresholds. ChemicalOCR substantially outperforms PaddleOCR v5 and EasyOCR for all IoU thresholds.

\begin{table*}
\centering
\resizebox{1.0\textwidth}{!}{%
\begin{tabular}{lcccccccccccccccc}
\toprule
\textbf{Models} & \multicolumn{4}{c}{\textbf{M2S (103)}} & \multicolumn{4}{c}{\textbf{USPTO-M (74)}}  & \multicolumn{4}{c}{\textbf{IP5-M (1000)}}\\
\cmidrule(lr){2-5} \cmidrule(lr){6-9} \cmidrule(lr){10-13} 
 & P & R & F1 & A & P & R & F1 & A & P & R & F1 & A \\
\midrule
\textit{@IoU $>$ 0.0} \\
PaddleOCR v5                     & 61.9 & 47.3 & 53.6 & 0.0  & 57.0  & 47.2  & 51.7  & 1.4  & 43.3 & 34.3 & 38.3  & 5.5  \\
EasyOCR                          & 10.4 & 11.4 & 10.8 & 0.0  & 29.8 & 17.0  & 21.7  & 0.0  &  30.6  & 19.7 & 24.0  & 3.5   \\
\textbf{ChemicalOCR (Ours)}      & \textbf{90.0}  & \textbf{90.6}  & \textbf{90.3} & \textbf{37.9} & \textbf{96.2} & \textbf{95.3} & \textbf{95.8} & \textbf{73.0} & \textbf{90.3} & \textbf{92.2}  & \textbf{91.3} & \textbf{73.7} \\
\midrule
\textit{@IoU $>$ 0.3} \\
PaddleOCR v5                    & 46.0   & 35.2  & 40.0  & 0.0  & 25.0 & 20.7  & 22.6  & 0.0  & 27.8  & 22.0  & 24.6   & 3.0  \\
EasyOCR                         & 10.2   & 11.2  &  10.7 & 0.0  & 29.0  & 16.6  & 21.1 & 0.0  & 29.6  & 19.1  & 23.3  & 3.2 \\
\textbf{ChemicalOCR (Ours)}     & \textbf{88.8}  & \textbf{89.4}  & \textbf{89.1}  & \textbf{40.0}  & \textbf{94.8}  & \textbf{94.0}  & \textbf{94.4} & \textbf{70.3} & \textbf{88.5} & \textbf{90.3}  & \textbf{89.4} & \textbf{72.8} \\
\midrule
\textit{@IoU $>$ 0.5} \\
PaddleOCR v5    & 8.9 & 6.8 & 7.7 & 0.0 & 2.3 & 1.1 & 1.2 & 0.0 & 2.2 & 1.7  & 1.9 & 0.6 \\
EasyOCR        & 9.8 & 10.7 & 10.2 & 0.0 & 24.8 & 14.2 & 18.0 & 0.0 & 23.5 & 15.2 & 18.4 & 2.7 \\
\textbf{ChemicalOCR (Ours)}     & \textbf{86.9} & \textbf{87.4} & \textbf{87.2} & \textbf{32.0} & \textbf{93.5} & \textbf{92.6} & \textbf{93.0} & \textbf{63.5} & \textbf{85.6} & \textbf{87.4} & \textbf{86.5} & \textbf{69.5} \\
\bottomrule
\end{tabular}
}
\caption{\textbf{Comparison of OCR performance for different IoU thresholds}: Comparison of our ChemicalOCR model with existing OCR models for different IoU thresholds.  The evaluation is conducted on real-world benchmarks (M2S, USPTO-M, and IP5-M). Precision $P$, Recall $R$, and F1 are measured at individual bounding-box level. Accuracy $A$ is measured at the image level; an image is considered correct if all OCR cells have an  IoU greater than the threshold (0.0, 0.3, 0.5), and their recognized characters match the ground truth.
}\label{abblation_ocr_scores}
\end{table*}

\section{Multi-Modal Models - Observations}

\begin{table}[!h]
\centering
\resizebox{1.0\columnwidth}{!}{%
\begin{tabular}{lcccc}
\toprule
\textbf{Models} & \multicolumn{2}{c}{\textbf{M2S}} & \multicolumn{2}{c}{\textbf{JPO}} \\
\cmidrule(lr){2-3} \cmidrule(lr){4-5}
 & Valid & Correct & Valid & Correct \\
\midrule
\textbf{DeepSeek-OCR}               & 15  & 0  & 72  & 32 \\
\textbf{GPT-5}                       & 30  & 3  & 74  & 19  \\
\textbf{MarkushGrapher-2}          & \textbf{95} & \textbf{56} & \textbf{92} & \textbf{71} \\
\bottomrule
\end{tabular}
}
\caption{\textbf{Comparison of (CX)SMILES predictions:} 
Comparison of the percentage of chemically-valid and correct (CX)SMILES generated by multi-purpose models (DeepSeek-OCR and GPT-5) versus our dedicated MarkushGrapher-2 model. 
“Valid” reports the percentage of chemically-valid (CX)SMILES, and “Correct” denotes the percentage of correct CXSMILES predictions (i.e., accuracy).}
\label{abblation_multi_purpose_models}
\end{table}

In Tables 2 and 3 of the main paper it was shown that our dedicated model MarkushGrapher-2 has substantially higher accuracy for molecular and Markush structure recognition than the multi-purpose models DeepSeek-OCR and GPT-5. Table \ref{abblation_multi_purpose_models} shows a more detailed comparison of the model predictions with respect to output validity and correctness. Specifically, we evaluate the proportion of chemically-valid (CX)SMILES versus correct (CX)SMILES given molecular and Markush structure images as input. The “Valid” metric captures only chemical validity of the model prediction and does not assess its correctness. The “Correct” metric reflects the accuracy of the predicted (CX)SMILES. We evaluate performance on two representative benchmarks, M2S for Markush structure recognition and JPO for molecular structure recognition.
The results reveal that both DeepSeek-OCR and GPT-5 predominantly generate chemically invalid outputs for Markush structure images (M2S benchmark). Typically, these invalid predictions constitute incorrect (CX)SMILES schemas, repetitive token loops, or unrelated text hallucinations. In contrast, the task of standard molecular structure recognition (JPO benchmark) leads to more chemically-valid outputs for both models. 
While DeepSeek-OCR is trained to predict SMILES as one of its subtasks \cite{wei2025deepseekocrcontextsopticalcompression}, it occasionally hallucinates random outputs when presented with Markush structure images. Figure \ref{fig:deepseek_outlier} shows an example in which DeepSeek-OCR generates a repeating sequence followed by the generation of an unrelated essay titled “How to Improve Your English Writing Skills.” For the datasets evaluated, GPT-5 did not exhibit this behavior. Rather, it occasionally reports its inability to generate CXSMILES from some Markush images. Figure \ref{fig:gpt_outlier} shows an example of such a prediction outlier.
The findings suggest that Markush images constitute a largely unseen image domain for the multi-purpose models DeepSeek-OCR and GPT-5, potentially lying outside their training data distribution. Our dedicated MarkushGrapher-2 substantially outperforms these multi-purpose models in both categories.

\section{Failure Analysis}

\begin{figure*}[!htbp]
    \centering
    \includegraphics[width=0.73\textwidth,
                     trim=0mm 60mm 90mm 0mm, 
                     clip]{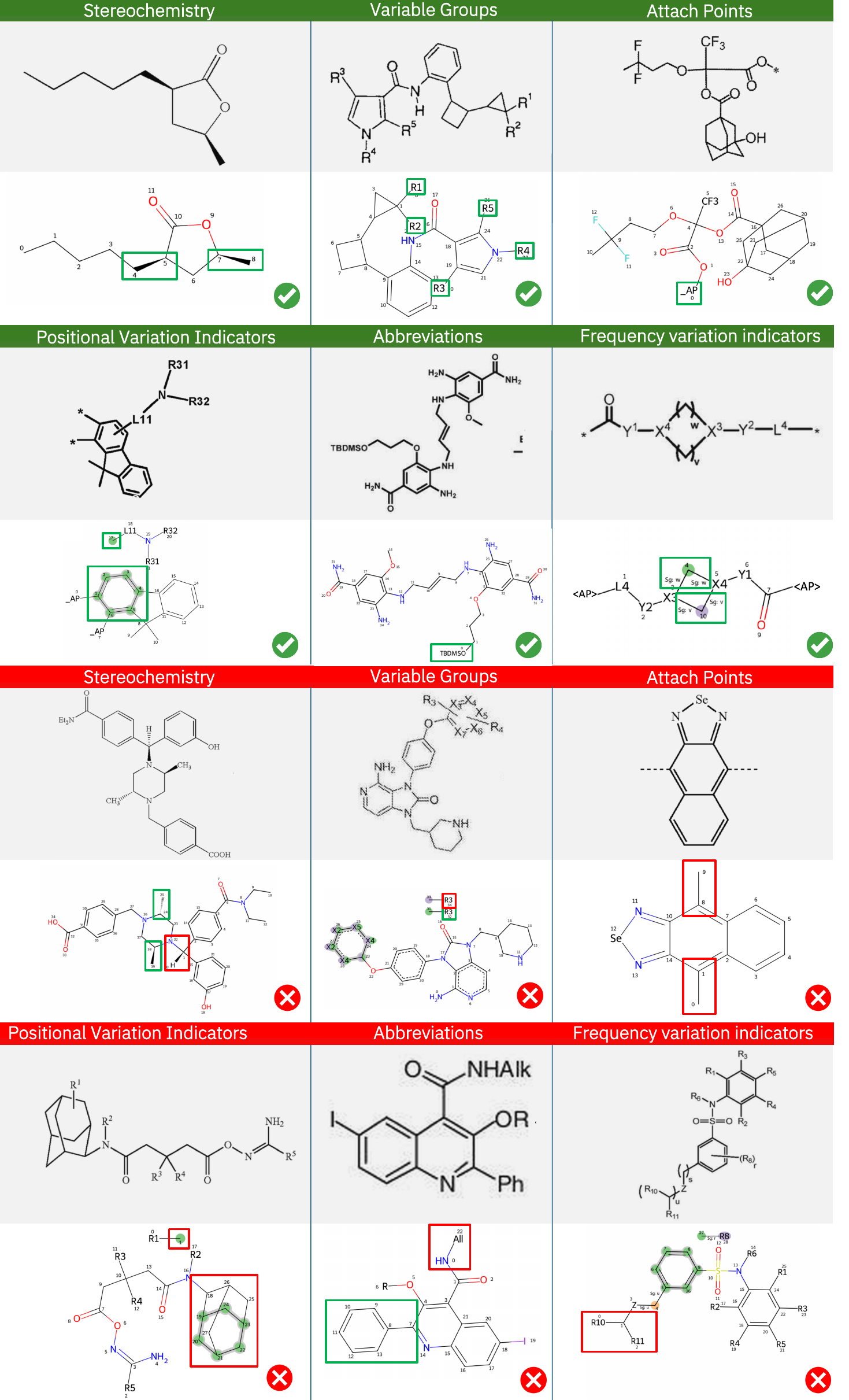}
    \caption{\textbf{Qualitative Evaluation}: MarkushGrapher-2 predictions of benchmark samples. A correct and an incorrect prediction is displayed for the features stereochemistry, variable groups, attach points, positional variation indicators, abbreviations, and frequency variation indicators.}
    \label{fig:qualitative_evaluation}
\end{figure*}

\autoref{tab:failure_modes} provides additional performance analysis on different Markush features for IP5-M, including stereochemistry. 
\autoref{fig:qualitative_evaluation} shows a qualitative evaluation of MarkushGrapher-2 predictions.
\begin{table}[h]
\centering
\resizebox{1.0\linewidth}{!}{
\begin{tabular}{lcccccc}
\hline
Methods & \multicolumn{6}{c}{IP5-M} \\ \cline{2-7} 
& \multicolumn{1}{c}{InChi} & \multicolumn{1}{c}{Variable} & \multicolumn{1}{c}{AP} & \multicolumn{1}{c}{m} & \multicolumn{1}{c}{Sg} & \multicolumn{1}{c}{A} \\ \hline
MarkushGrapher-2  &   73.3(70.8) & 74.8 & 73.9 & 78.8 & 30.7 & 53.7(51.5)  \\ 
\end{tabular}}
\caption{\textbf{Comparison of Markush features performances}: Breakdown of model performance for backbone structure (InChi), variable groups, attach points, positional and frequency variation indicators and total accuracy. Scores reported in parentheses consider stereochemistry. (6.4\% of samples contain stereochemistry).}
\label{tab:failure_modes}
\end{table}
\newline
\newline
\newline
\newline
\newline
\newline
\newline
\newline
\newline

\begin{table*}
\centering
\resizebox{\textwidth}{!}{%
\begin{tabular}{lccccccccc}
\toprule
\textbf{Benchmarks} & Samples &\multicolumn{4}{c}{Percentage of CXSMILES with at least one} & Mean num. atoms & Mean num. OCR cells\\
\cmidrule(lr){3-6} 
 & & Variable group & Attach point & m-section & Sg-section &  &   &   \\
\midrule
M2S        & 103   & 97  & 0 & 30 & 25  & 19  & 15      \\
USPTO-M    &  74   & 91  & 0 & 74 & 42& 20  & 8      \\
WildMol-M  & 10000 & 59 & 31 & 15  & 12& 14  & -       \\
IP5-M      & 1000  & 60  & 28 & 17 & 26  & 14 &  7      \\
\bottomrule
\end{tabular}
}
\caption{\textbf{Benchmarks - Statistics}: Comparison of the number of samples, percentage of dataset samples containing at least one Markush feature (variable group, attach point, positional variation indicator (m-section), frequency variation indicator (Sg-section)), mean number of heavy atoms, and mean number of OCR cells.
}\label{abblation_bechmark_statistics}
\end{table*}

\noindent
\begin{minipage}[t]{0.50\textwidth}
    \centering
    \includegraphics[width=\linewidth,
                     trim=8mm 250mm 218mm 2mm, clip]{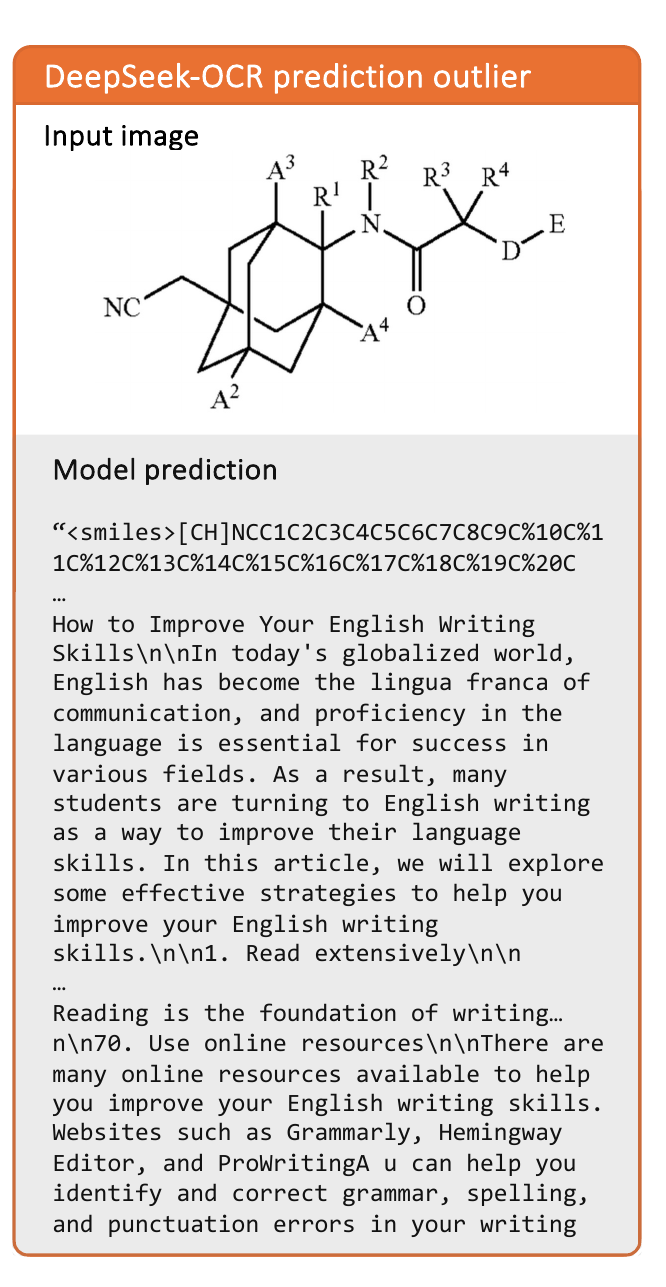} 
    \captionof{figure}{\textbf{DeepSeek-OCR}: Example of a model prediction outlier.}
    \label{fig:deepseek_outlier}
\end{minipage}%
\hfill
\begin{minipage}[t]{0.50\textwidth}
    \centering
    \includegraphics[width=\linewidth,
                     trim=8mm 250mm 218mm 2mm, clip]{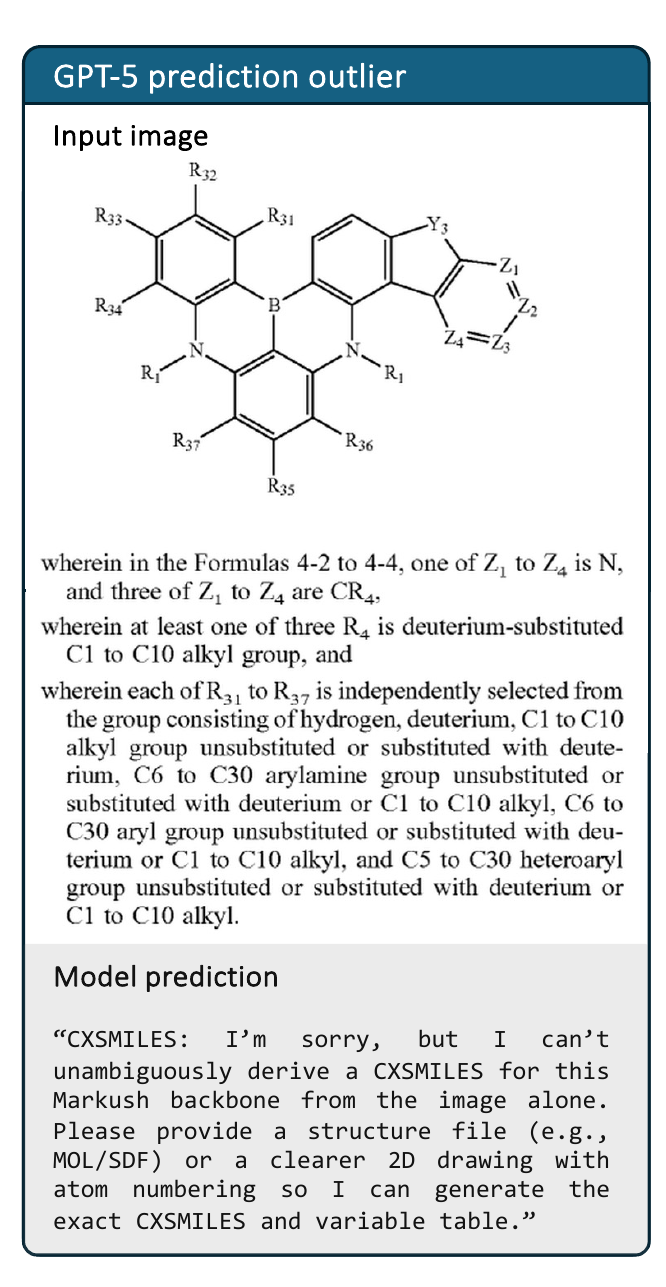}
    \captionof{figure}{\textbf{GPT-5}: Example of a model prediction outlier.}
    \label{fig:gpt_outlier}
\end{minipage}

%

{
    \small
    \bibliographystyle{ieeenat_fullname}
    \bibliography{main}
}


\end{document}